\documentclass{article}

\usepackage{arxiv}

\usepackage[utf8]{inputenc} 
\usepackage[T1]{fontenc}    
\usepackage{hyperref}       
\usepackage{url}            
\usepackage{booktabs}       
\usepackage{amsfonts}       
\usepackage{nicefrac}       
\usepackage{microtype}      
\usepackage{lipsum}		
\usepackage{graphicx}
\usepackage{natbib}
\usepackage{doi}

\usepackage{graphicx}
\usepackage{amsmath}
\usepackage{amssymb}
\usepackage{multirow}
\usepackage{longtable}
\usepackage{xcolor}
\usepackage{subfig}

\usepackage{algorithm} 
\usepackage{algpseudocode}

\algnewcommand\algorithmicforeach{\textbf{for each}}
\algdef{S}[FOR]{ForEach}[1]{\algorithmicforeach\ #1\ \algorithmicdo}
\algdef{SE}[SUBALG]{Indent}{EndIndent}{}{\algorithmicend\ }%
\algtext*{Indent}
\algtext*{EndIndent}

\title{Towards Counterfactual and Contrastive Explainability and Transparency of DCNN Image Classifiers}

\date{} 					

\author{ Syed Ali Tariq \\
	Department of Computer Science\\
	COMSATS University Islamabad\\
	Islamabad, Pakistan \\
	\texttt{s.alitariq1@gmail.com} \\
	\And
	Tehseen Zia \\
	Department of Computer Science\\
	COMSATS University Islamabad\\
	Islamabad, Pakistan \\
	\texttt{tehseen.zia@comsats.edu.pk} \\
	\AND
	Mubeen Ghafoor \\
	School of Computer Science (SoCS) \\
	University of Lincoln \\
	Lincoln, UK \\
	\texttt{mubeenghafoor@yahoo.com} \\
}

\begin{document}
\maketitle

\begin{abstract}
	Explainability of deep convolutional neural networks (DCNNs) is an important research topic that tries to uncover the reasons behind a DCNN model's decisions and improve their understanding and reliability in high-risk environments.
	In this regard, we propose a novel method for generating interpretable counterfactual and contrastive explanations for DCNN models. The proposed method is model intrusive that probes the internal workings of a DCNN instead of altering the input image to generate explanations. Given an input image, we provide contrastive explanations by identifying the most important filters in the DCNN representing features and concepts that separate the model's decision between classifying the image to the original inferred class or some other specified alter class. On the other hand, we provide counterfactual explanations by specifying the minimal changes necessary in such filters so that a contrastive output is obtained.
	Using these identified filters and concepts, our method can provide contrastive and counterfactual reasons behind a model's decisions and makes the model more transparent. One of the interesting applications of this method is misclassification analysis, where we compare the identified concepts from a particular input image and compare them with class-specific concepts to establish the validity of the model's decisions. The proposed method is compared with state-of-the-art and evaluated on the Caltech-UCSD Birds (CUB) 2011 dataset to show the usefulness of the explanations provided.
\end{abstract}


\section{Introduction}

In recent years, deep convolutional neural networks (DCNNs) have achieved state-of-the-art performance in many computer vision applications such as medical imaging and diagnostics \cite{gu2019net, shakeel2019lung}, biometrics identification \cite{liu2019adaptiveface}, 
object detection \cite{wang2017fast}, scene segmentation \cite{fu2019dual}, image inpainting \cite{yu2019free}, etc.  DCNNs are a type of deep learning (DL) method that automatically learns generalized representations from the input data useful for image classification, segmentation, or recognition tasks. 
Most of the current research on DCNNs is focused on architectural improvements \cite{he2016deep, chen2017dual,tan2019efficientnet, touvron2019fixing, zhang2020resnest,mohan2020efficientps}. Although this aspect of research is necessary, one essential aspect that is mostly overlooked, missing, or not focused upon while developing new CNNs is their explainability. Since DCNNs are trained in an end-to-end manner, their inner workings are not well understood, which makes them a black-box \cite{arrieta2020explainable}. 

\textcolor{black}{The explainability or transparency of DCNN models is extremely important for certain applications where the impact of incorrect or unjustifiable predictions can have a significant impact on the outcome \cite{samek2017explainable, goebel2018explainable, rudin2019stop}.} \textcolor{black}{This is especially true for those security-critical applications that involve endangerment of human life or property, such as in medical imaging and diagnostics \cite{tjoa2020survey,holzinger2017we}, autonomous vehicles \cite{zablocki2021explainability}, military applications \cite{svenmarck2018possibilities}, etc.} 
If a DCNN model's decision cannot be interpreted or it cannot explain how or why it made its decision, then using such a model in high-risk environments will carry some risks. 
\textcolor{black}{For example, DCNNs are known to be affected by dataset bias \cite{zhang2018examining} due to which they may rely on unrelated or out-of-context patterns to classify images. Zhang \textit{et al.} \cite{zhang2018examining} demonstrated an example where a CNN incorrectly relied on eye features to identify the `lipstick' attribute of a face image. Similarly, if a DCNN model is trained to detect lung diseases from chest X-rays annotated with doctors' pen marks, it may learn to rely on them to make predictions. Another issue with DCNNs is that they are prone to adversarial attacks where subtle changes to the input may lead to the CNN producing incorrect results \cite{akhtar2018threat}. Adversarial attacks pose threats to many security-critical applications, such as for self-driving cars where minor obstructions on traffic signs can cause incorrect decisions \cite{eykholt2018robust} or in surveillance systems where malicious users can cause harm \cite{thys2019fooling}. These are some of the many aspects that make DCNNs un-trustworthy and require explainable AI techniques to identify their weaknesses and train robust, trustworthy, and transparent models \cite{rudin2019stop, ghorbani2020neuron, fong2017interpretable}}.

In the literature, several types of explainability or interpretability methods exist. At the top level, most types of explainable techniques can be broadly divided into two main categories: intrinsic/inherent interpretability or post-hoc interpretability \cite{du2019techniques}. Inherently interpretable techniques attempt to introduce explainability built into the structure of the model itself that makes the model self-explanatory, for example, decision trees. \textcolor{black}{Whereas post-hoc interpretable techniques usually consist of a separate explanation of a pre-trained DCNN black-box model. These techniques may use a second model to explain the black-box either by probing the internal structure of the black-box or by altering the input.} An example of post-hoc explainability is a visual explanation technique such as GradCAM \cite{selvaraju2017grad}. 
Counterfactual and contrastive explanations are two popular types of post-hoc explanation methods. 
Contrastive explanations generally try to find the critical features in the input that lead to the model making its decision to the inferred class \cite{dhurandhar2018explanations}.
While in counterfactual explanations, the goal is to alter the input features (pixels) such that the model changes its decision to some other counter class \cite{pmlr-v97-goyal19a}.
\textcolor{black}{Such explanations are natural to humans since they mimic human thought processes.} For example, a contrastive explanation can have the form \textit{``if some event X had not occurred, then event Y would not have occurred"}. Whereas using counterfactuals, we can ask questions such as \textit{``if X is changed to X' then would Y' happen instead of Y?"}. Such types of explanations are human-friendly and easy to understand. \textcolor{black}{If a DCNN model can provide such explanations, it can be considered a reliable or trustworthy model, and we can predict its behavior.} 

Recently, several counterfactual and contrastive explanation methods have been proposed \cite{pmlr-v97-goyal19a, hendricks2018grounding, dhurandhar2018explanations, liu2019generative, luss2019generating}. 
These methods generally perturb the input pixels to alter the model prediction. One of the drawbacks of such an approach is that it generally does not identify semantically meaningful features or high-level concepts useful in explaining the model decisions. 
\textcolor{black}{Another issue with pixel-based perturbation methods is that the search space for finding the optimal combination of pixels that can cause the network to change its prediction is large, which makes such methods computationally expensive \cite{wang2020scout}.} 
A better approach would be to identify the major concepts that a model learns and relies on to make predictions.
In a recent study, Akula \textit{et al.} \cite{akula2020cocox} tried to address this issue by using super-pixels to identify critical concepts that, when added or removed from the input image, alter the model's decision. \textcolor{black}{Although this method provides a useful way to generate explanations, it still operates on pixel data to generate explanations that do not make it fully transparent.} 
In another work, concept activation vectors (CAV) \cite{pmlr-v80-kim18d} were introduced that essentially measure the sensitivity of the model towards a particular high-level concept. \textcolor{black}{The concepts can be learned from either training or user-provided data.}
\textcolor{black}{However, this method only identifies whether a particular concept is important or not. It does not investigate if it is actively used in the decision-making process for a particular input.}

The problem with these works is that they are primarily non-intrusive. They only look at model behavior by altering the input and do not thoroughly investigate the internal workings or reasoning behind the model predictions. Such interpretability methods may not be aligned with how the model is making decisions. 
In this study, we propose a post-hoc explainability method that predictively identifies counterfactual and contrastive filters in a DCNN model. Although in this work, we restrict the identification of filters from the top convolution layer only, the proposed method can be used to identify counterfactual and contrastive filters from any layer of a DCNN and can also be used to identify such filters in other networks. 
It has been shown that filters in the top convolution layer of a DCNN tend to learn more abstract, high-level features, concepts, and even whole objects as well  \cite{bau2020understanding, zhou2014object, bau2017network}.
We show that when enabled, disabled, or modified in a certain way, these filters can make the DCNN predict the input image either to the original inferred class or to some chosen alter class. 
\textcolor{black}{These identified filters represent the critical concepts and features that the model learns and are chosen to maximize the model's prediction towards the specified class.} Essentially, the proposed method identifies the most important features or concepts that separate the model's decisions between the two classes.
\textcolor{black}{In this regard, the proposed solution is a predictive counterfactual explanation (CFE) model trained on top of a pre-trained DCNN model to generate explanations.
The proposed method has the following three main objectives:} 
\begin{enumerate}
    \item \textcolor{black}{Identify the minimum set of filters necessary to predict the input image to the inferred class. We call these filters minimum correct (MC) for the input image with respect to its inferred class.
    \item \textcolor{black}{Identify the minimum set of filters that if they were additionally activated, 
    it would have altered the model's decision to some counter class $c'$. We call this set of filters minimum incorrect (MI) for the input image with respect to its counter class $c'$.}} 
    \item \textcolor{black}{Using the identified MC and MI filters, provide contrastive and counterfactual explanations by highlighting the presence or absence of the features and concepts represented by these filters and demonstrate how they affect the model's decisions.}
\end{enumerate}

The primary motivation behind this work is to provide clear and easy-to-understand explanations that do not require a domain expert to use the system. The need for transparent DCNN models in high-risk environments is also a crucial factor. 
\textcolor{black}{We believe that explanations provided by the proposed method could be useful for applications such as machine teaching or model debugging. We can utilize the MC or MI filters (by visualizing their receptive fields) to teach human users what features are essential for the inferred class or some alter class, respectively. Such explanations can help humans differentiate between similar-looking classes under challenging tasks such as fine-grained classification of birds or medical imaging diagnostic tasks such as brain tumor detection. The other application that can benefit from the proposed method is model debugging. Expert users can use MC or MI filters identified by the proposed method to detect weak or faulty filters that may cause dataset bias or misclassifications. Such filters can be disabled permanently or retrained in a controlled manner to repair the model. This approach can also be used to detect and prevent adversarial attacks and establish a ‘trust index’ for the DCNN model's predictions.}

\begin{figure*}[t!]
\begin{center}
    \includegraphics[width=1.0\textwidth]{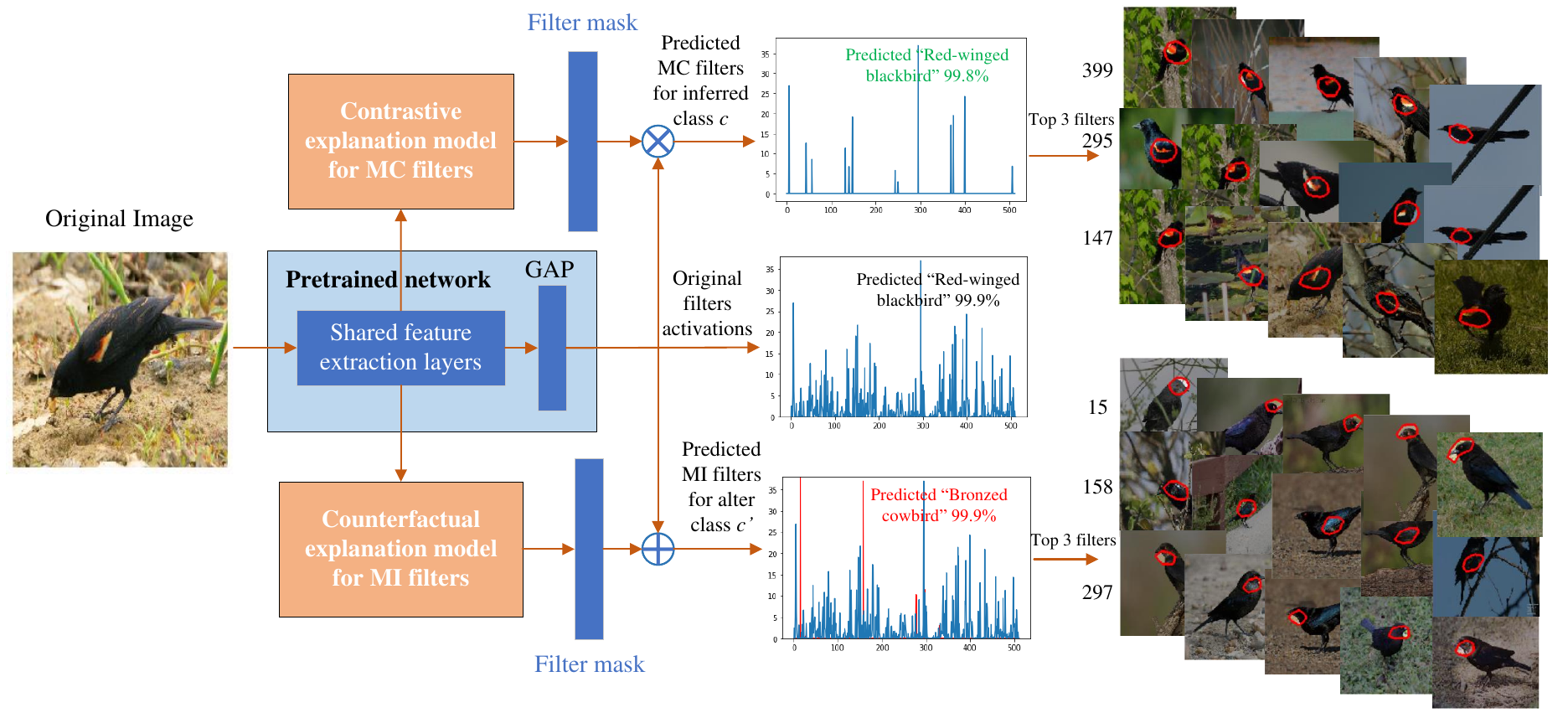}
\end{center}
  \caption{\textcolor{black}{
  Explanation provided by the proposed approach. Our method identifies the most important minimum correct (MC) and minimum incorrect (MI) filters using which the pre-trained model either predicts the input image to its original inferred class or some chosen alter class, respectively. \textcolor{black}{The top-3 MC filters for the example image classified as ``Red-winged blackbird" show that the red spot on this bird's wing is the most discriminating feature for it. Whereas, the top-3 MI filters} of this example for the ``Bronzed cowbird" class show that if the filters corresponding to features such as bird's red eyes and the blue-tinged feather were present in the input, the model would have been more likely to predict this image as ``Bronzed cowbird."}} 
\label{fig:top_fig}
\end{figure*}

\textcolor{black}{An example explanation provided by the proposed method is shown in Fig. \ref{fig:top_fig}. 
\textcolor{black}{In Fig. \ref{fig:top_fig}, our approach provides a contrastive explanation by identifying the most critical MC filters using which the model maintains the prediction to the original inferred class, i.e. ``Red-winged Blackbird". 
Activation magnitudes of the 13 MC filters predicted by the CFE model are shown as a graph for the given input in Fig. \ref{fig:top_fig}.} Using just these filters, the VGG-16 model still classifies the input to its original inferred class with 99.8\% confidence. Visualization of the top-3 filters on different images of the inferred class shows the most important features associated with the respective filters for this class. 
\textcolor{black}{
The counterfactual explanation in Fig. \ref{fig:top_fig} identifies the MI filters for some chosen alter class (usually top-2 or 3 class), which in this case is the ``Bronzed cowbird".} The CFE model predicts the minimum additive values to alter the activation magnitude of these filters such that the input image is classified to the alter class, i.e. ``Bronzed cowbird". \textcolor{black}{The graph for MI filters in Fig. \ref{fig:top_fig} 
shows the modified activation magnitudes highlighted in red.} Top-3 MI filters show the most important features for this alter class. Filter 15 activates the most on the red-colored eye of the Bronzed cowbird, while filter 158 activates the most on the blueish tinge on the bird's wings. These two filters identify the critical features that if they were present in the input image, the model would have been more likely to classify the bird as ``Bronzed cowbird" instead of the ``Red-winged blackbird" class. We can say that these features separate the model's decision between the two classes.} 

In providing such types of explanations, 
the proposed method probes the internal working of a DCNN model. This makes the provided explanations more meaningful, trustworthy and provides insights into how a trained model makes its decisions, thus making it more transparent.
\textcolor{black}{To the best of our knowledge, this is the first study that has investigated the importance of concepts learned by individual filters in terms of providing both counterfactual and contrastive explanations and how they contribute to the network's overall prediction.}
The rest of the paper is structured as follows. Section \ref{sec:two} discusses the related works. Sections \ref{sec:three} presents the proposed approach. Section \ref{sec:four} presents the results and discussion, and Section \ref{sec:five} concludes the paper.


\section{Related Work}
\label{sec:two}


Since DCNNs are trained end-to-end, they learn complex hidden representations in their intermediate and end layers to map the input data with the respective target outputs. To understand what the network has learned, many post-hoc visual explanation methods in literature try to investigate the relevance or importance between the input features (i.e., pixels of the image) and the model's predictions. These methods can be generalized into different categories such as back-propagation based methods \cite{simonyan2013deep, bach2015pixel}, activation based methods \cite{zhou2016learning, selvaraju2017grad}, or perturbation based methods \cite{ribeiro2016should, fong2017interpretable, petsiuk2018rise}.
Back-propagation-based methods try to determine the importance of different input features/pixels by back-propagating the error in the output back towards the input image. An example of such a method is layer-wise relevance propagation (LRP) \cite{bach2015pixel}. Activation-based methods provide visual explanations by generating heatmaps that identify the regions in the input image that were useful for classifying images to some target class. 
One of the most popular works 
for this type of explainability is the class activation map (CAM) \cite{zhou2016learning} and its variant Grad-CAM \cite{selvaraju2017grad}.
Perturbation-based explanation methods 
involve modifying the input of the model by either preservation or deletion of different portions/pixels of the input image and observing how it affects the model's output \cite{wagner2019interpretable}. 

The main drawback of such visual explainability methods is that they are primarily helpful in explaining the prediction with respect to the input. They are not suitable for understanding why or how the model has made that decision. Such explanations can often be misleading \cite{rudin2019stop}. \textcolor{black}{For example, if a model makes an incorrect decision, then the visual explainability method would still try to visualize the input region that is found important in making the incorrect decision. These methods do not explore the internal workings of the model, which remains a black-box.}

Some of the shortcomings of visual explainability methods can be addressed by intrinsic or inherently interpretable methods. 
These models provide more useful interpretations than visual explanations, making them suitable for high-stakes applications. 
In recent years, researchers have developed inherently interpretable DL models for different computer vision applications that have produced promising results \cite{liu2019tabby,zhang2018interpretable,zhang2019interpreting,chen2019explaining, chen2019looks}. 
Liu \textit{et al.} \cite{liu2019tabby} proposed an interpretable object classification model that learns classification criteria based on hierarchical visual attributes that define the category or class of an image. The authors devised a two-stream architecture, where the first stream is a conventional CNN structure that learns image attributes from the training images. At the same time, the second stream takes as input the hierarchical category labels that define the input image and learns a linear combination of different attributes that define each category. The outputs from both streams are classified jointly to learn the hierarchical attribute structure present in images. One main drawback of this approach is that hierarchical category labels have to be provided for training which may not be readily available.
Chen \textit{et al.} \cite{chen2019looks} developed a very useful interpretable model for image recognition applications based on network dissection \cite{bau2017network}. The proposed model inspects different parts of the input image, which the authors call prototypes, and finds the nearest similar prototypical parts learned from the training dataset. The prototypical parts for different classes are learned by clustering together the semantically similar image patches encountered during training. The authors have demonstrated that their interpretable model can achieve similar accuracy compared to standard CNN models while providing useful explanations.
In another work, Zhang \textit{et al.} \cite{zhang2018interpretable}, modified the top convolutional layers of standard CNN models to make them interpretable. The authors did so by introducing a loss for each filter that pushes it to learn some specific object part belonging to some specific image category. When the model is fully trained, the representations learned by the interpretable filters are disentangled. 
\textcolor{black}{This property of interpretable filters is advantageous in providing relevant explanations and measuring the contribution of different object parts on the model's prediction \cite{chen2019explaining}}. 


Another promising area of research for interpretable models is network de-coupling \cite{li2019dynamic, hu2020architecture, liang2020training}. Such approaches aim to understand and manipulate the inner workings of the CNN model to make them interpretable. Li \textit{et al.} \cite{li2019dynamic} proposed a method to de-couple network architecture by dynamically selecting suitable filters in each layer that form a hierarchical calculation path through the network for each input image. 
\textcolor{black}{Different filters selected in each layer form a de-coupled sub-architecture that corresponds to different semantic concepts.} 
In another study, Saralajew \textit{et al.} \cite{saralajew2019classification} proposed a classification-by-components network that extracts components (i.e., visual features that are representative of one or more classes) from input images and reasons over them describing which components are either positively or negatively related to a particular class and which components are not important at all.
Recently, Ghorbani \textit{et al.} \cite{ghorbani2020neuron} proposed a method to quantify the contribution of each neuron or filter in a deep CNN towards the model's accuracy. The authors found that there is usually a small number of critical neurons that, if removed from the network, vastly decrease the overall performance of the DCNN.

\textcolor{black}{Although the interpretable DL models such as the ones discussed here help provide trustworthy explanations, they still have some drawbacks.} These methods usually devise complex mechanisms to ensure interpretability, and developing them is significantly more challenging. Furthermore, these methods need deeper analysis to understand or decipher the explanations that may require domain expertise to implement and use, making them inaccessible for non-expert users. For example, the techniques used to implement the ProtoPNet model \cite{chen2019looks} are not yet available in the standard DL libraries used for training CNNs \cite{rudin2019stop}. This makes designing such networks costly, time-consuming, and problematic.
Additionally, since inherently interpretable models must satisfy additional constraints to satisfy interpretability, they usually demonstrate lower accuracy as compared to standard black-box models \cite{zhang2018interpretable,liu2019tabby}. 

On the other hand, counterfactual explanation methods attempt to provide more human-friendly explanations and are easy to understand. Goyal \textit{et al.} \cite{pmlr-v97-goyal19a} generated counterfactual explanations by identifying regions in the input image that can be changed such that the network's decision is altered from the original class to some specified counter class. Similarly, Wang \textit{et al.} \cite{wang2020scout} proposed a counterfactual explanation method that generates an attributive heatmap that is indicative of the predicted class but not of some counter class. In another work, Akula \textit{et al.} \cite{akula2020cocox} proposed counterfactual explanations based on semantic concepts that can be added or removed from the input image to alter the model decisions. \textcolor{black}{Although these methods help provide understandable and user-friendly explanations,} they do not explore the internal workings of the network and do not make them transparent. Additionally, Goyal \textit{et al.} \cite{pmlr-v97-goyal19a} perform an exhaustive search for pixels and features to modify that alter model prediction. Such a method can become too complex and slow to generate explanations.

In this study, we tried to address some of the existing issues with DCNN interpretability methods to propose an explainability method that generates contrastive and counterfactual explanations by probing internal network structure to make them more transparent. The proposed method identifies the minimum number of most essential filters from the top convolution layer of a pre-trained DCNN for a particular input image that, when enabled, disabled, or modified, alters the model's decision to some specified class. Our work is similar to the classification-by-components \cite{saralajew2019classification} approach in the sense that instead of components, we identify the crucial filters corresponding to high-level concepts that the pre-trained DCNN model learns and relies on for the classification of images to a particular class. The proposed work is more straightforward and provides contrastive, easy-to-understand explanations that are natural to humans as compared to network de-coupling approaches \cite{li2019dynamic, hu2020architecture}. The proposed work is also better than existing counterfactual explanation approaches \cite{akula2020cocox, pmlr-v97-goyal19a, hendricks2018grounding} that operate on pixel data to generate explanations. In contrast, we probe the internal filters to generate more meaningful explanations that make the network more transparent.

\begin{figure*}[hbt!]
\begin{center}
    \includegraphics[width=1.0\linewidth]{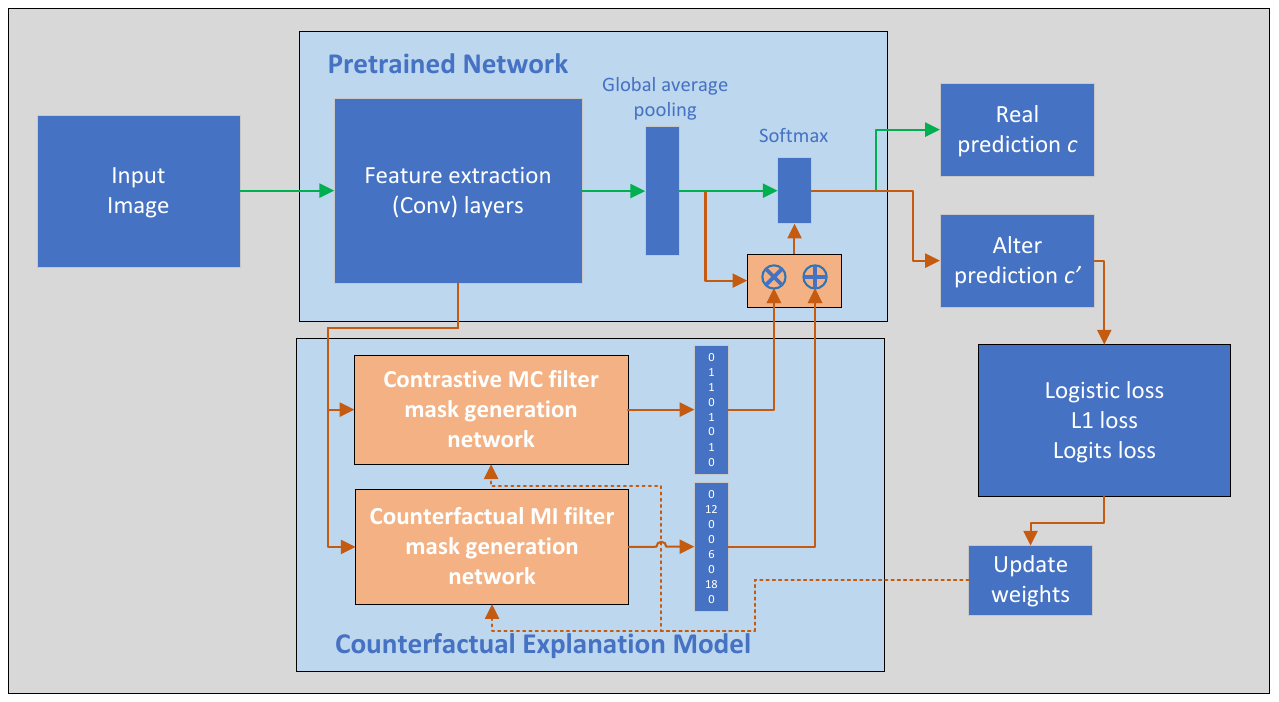}
\end{center}
   \caption{Overall block diagram of the proposed counterfactual and contrastive explanation model. Given an input image, the contrastive and counterfactual filter generation networks predict the MC and MI filter maps. MC filters are multiplied with the pre-classification output of the pre-trained network to disable all but the important features using which the model is able to maintain prediction to the original inferred class. Similarly, MI filter map is used to alter the activation magnitudes of the pre-trained model such that the model predicts the image to some alter class $c'$.}
\label{fig:block_daigram}
\end{figure*}

\section{Proposed methodology}
\label{sec:three}
Given a pre-trained DCNN model, $M$, and the dataset, $D$, consisting of images belonging to $C$ distinct classes, our objective is to provide two types of explanations for each image $x_i \in D$. Firstly, our model predicts the minimum set of filters in the top convolution layer of $M$ necessary for $M$ to maintain its prediction of image $x_i$ to the original inferred class (source class) $c_i \in C$. \textcolor{black}{We call these filters as minimum correct (MC) for image $x_i$ to be classified to class $c_i$, denoted as $F_{MC_i} \in [0,1]^{1\times n}$,
where $n$ is the number of filters in the top convolution layer of $M$. In the VGG-16 DCNN model, the number of filters in the top convolution layer is 512, i.e., $n=512$.} Values of `1' and `0' indicate whether the corresponding filter is predicted to be active or disabled, respectively. 

Secondly, our model predicts the minimum set of filters that, if they were altered by a larger magnitude, would have resulted in the DCNN classifying the input to some target class $c'_i \in C$. \textcolor{black}{We call these filters as minimum incorrect (MI) for image $x_i$ with respect to the target class $c'_i$, denoted as $F_{MI_i} \in [\mathbb{R}^+]^{1\times n}$.} 
Non-zero indexes in $F_{MI_i}$ correspond to the MI filters, and the values at these indexes indicate the magnitude by which the original filter activations are altered to modify the DCNN's decision. The overall diagram depicting the proposed approach is shown in Fig. \ref{fig:block_daigram} which is described in the following sub-sections. 

\subsection{CFE model for MC filters}
\textcolor{black}{To achieve the first objective, we train the CFE model to predict MC filters from the pre-trained model $M$ that explain each image with respect to its inferred class.} 
\textcolor{black}{The MC CFE model is designed with partially similar architecture as $M$ by sharing the feature extraction layers up to the top convolution layer of $M$, as shown in Fig. \ref{fig:block_daigram}. 
\textcolor{black}{Given an input image $x_i$, model $M$ generates two outputs: (1) the feature maps $g_i \in [\mathbb{R}^+]^{1\times n}$ produced at the last convolution layer of $M$ after the global average pooling layer, and (2) the source or inferred class $c_i$.
The MC CFE model takes as input the feature maps $g_i$ and generates a binary filter map $F_{MC_i}$ corresponding to class $c_i$. With these definitions, the MC CFE model can be represented as:}
}
\begin{equation}\label{eq:new4}
\begin{aligned}
    \textcolor{black}{F_{MC_i}} & \textcolor{black}{= CFE_{MC}(g_i)} \\
    & \textcolor{black}{ = ReLU_t(A(d^n(g_i))),}
\end{aligned}
\end{equation}
\textcolor{black}{
\textcolor{black}{where $g_i$ represents the feature maps after the global average pooling layer of $M(x_i)$,} $d^n$ is a dense layer with $n$ units, $A$ represents sigmoid activation function, and $ReLU_t$ is thresholded-ReLU layer with threshold $t$ set to $t = 0.5$ that outputs the approximately\footnote{\textcolor{black}{$F_{MC_i}$ is not exactly binary but it is made close to binary as training progresses. We fully binarize it at inference time to ensure that no un-intended scaling is taking place on the activation magnitudes. If we hard binarize it during training using a fixed threshold, then the function does not remain differentiable and model cannot learn weights.}} binarized MC filter map $F_{MC_i}$. The thresholded-ReLU function sets all values below the threshold to zero and keeps other values unchanged. 
The shared feature extraction layers between the CFE model and the pre-trained model $M$ are kept frozen for the training of the CFE model. Additionally, all layers of $M$ after the top convolution layer are frozen. \textcolor{black}{Only the dense layer $d^n$ weights are updated during training of the CFE model.
To generate explanations, $CFE_{MC}$ model predicts the MC filters matrix for the corresponding input $x_i$.}
}
\textcolor{black}{$F_{MC_i}$ is multiplied with $g_i$ (Hadamard product) to disable all but the MC filters from the top layer of $M$.} The DCNN model $M$ makes the alter prediction with the disabled filters as $\hat{c_i}$:   
\begin{equation}\label{eq:new3}
    \textcolor{black}{\hat{c_i} = h( g_i \circ F_{MC_i}),}
\end{equation}
\textcolor{black}{where $h$ represents the classification (fully-connected and softmax) layers of $M$.}

\textcolor{black}{
The MC CFE model is trained to predict the optimal filter maps $F_{MC_i}$ to reduce the loss between the predicted classes $\hat{c}_i$ and the source class $c_i$. The optimal MC filters $F_{MC_i}$ are learned by minimizing the following three losses simultaneously: 1) cross-entropy (CE) loss, $L_{CE}$, for classifying each input image $x_i$ by modified model $M$ to the specified class $c$, 2) sparsity loss, $L_{l1}$, that ensures $F_{MC_i}$ is sparse so that minimal filters remain active, and 3) negative logits loss, $-L_{logits}$, that ensures that the predicted sparse filters have higher contribution towards the chosen class $c$:
\begin{equation}\label{eq:1}
    L_{MC} = L_{CE}(\hat{c}_{i}, c_{i}) + \lambda L_{l1}(F_{MC_i}) - L_{logits},
\end{equation}
where $\lambda$ is the weight assigned to the sparsity loss. $L_{CE}(\hat{c}_{i}, c_i)$ is computed using the output of the modified model $M$ and the desired class $c_i$ for each training example $i$:
\begin{equation}\label{eq:2}
    L_{CE}(\hat{c}_{i}, c_i) = -\frac{1}{m}\sum_{i=1}^{m} [c_i\log \hat{c}_{i} + (1-c_i)\log (1-\hat{c}_{i}))].
 \end{equation}}
\textcolor{black}{The $L_{l1}(F_{MC_i})$ loss minimizes the sum of the activated filters that push the} \textcolor{black}{CFE model to predict minimally sufficient filters:} 
\begin{equation}\label{eq:3}
    L_{l1}(F_{MC_i}) = \sum_{i=1}^{m} \sum_{k=1}^{n}||F_{MC_i}(k)||,
 \end{equation}
where $n$ is the number of filters in the top convolution layer of $M$.

The negative logits loss $-L_{logits}$ is necessary for the proposed methodology to ensure that the sparse filters predicted by the MC CFE model are contributing maximally towards the source class $c_i$. This loss is applied on the logits (weighted sum) computed after disabling the filters using $F_{MC_i}$ and before applying the final activation function to get $M$'s output:
\begin{equation}
    L_{logits}(F_{MC_i}) = -\sum_{k=1}^{n}||(F_{MC_i}(k) * \textcolor{black}{g_i(k))} * W_{k,c_i}||,
 \end{equation}
\textcolor{black}{where $g_i$ represents the feature maps after the global average pooling layer of model $M$}, and $W_{k,c}$ represents the pre-trained weights of model $M$ connecting the GAP layer with the output layer's class $c_i$.
The negative sign of the loss is for ensuring that the model chooses filters that have higher weight for the desired class, and hence their activation contributes more towards it and results in a larger logits score. The results section shows that if this loss is not included, the CFE model is more likely to predict less important filters that may still classify the images to the desired class but with lower confidence. 

When all three of these losses are minimized, the CFE model learns to predict minimally sufficient or MC sparse filters using which the inputs are classified to the source class $c_i$. For example, if the pre-trained model $M$ classified a given image $x_i$ to class $c_i$, then to identify the MC filters, we train the CFE model with respect to the source class $c_i$. This CFE model is then used to predict the most important $F_{MC_i}$ filters necessary for classifying input $x_i$ to class $c_i$.
\textcolor{black}{The procedure for MC CFE model training is summarized in Algorithm \ref{algo:1}}

\begin{algorithm}
	\caption{Procedure for training MC CFE model
	} 
	\label{algo:1}
	\textbf{Input: }{Image $I$, DCNN model $M$, target class $c$, dataset $D$}
	\begin{algorithmic}[0]
	    \State 1. Train MC CFE model for target class $c$ over the training dataset $D$
	    \Indent
	    \ForEach{image $x \in D$}
    	    \State \textcolor{black}{$g,c = M(x)$} 
    	    \State $F_{MC} = \textcolor{black}{ReLU_t(A(d^n(g)))}$ 
    	    \State $\hat{c} = h( g \circ F_{MC})$ \Comment{alter prediction with just MC filters enabled}
    	    \State{Minimize MC loss Equation \ref{eq:1}}
	        \Indent
    	        \State {$L_{MC} = L_{CE}(\hat{c}, c) + \lambda L_{l1}(F_{MC}) - L_{logits}$}    
    	   \EndIndent
	    \EndFor
	    \EndIndent
	    \State 2. Generate contrastive explanation using MC CFE model for input image $I$
	    \Indent
    	    \State $g,c = M(I)$ 
    	    \State $F_{MC} = ReLU_t(A(d^n(g)))$ 
        \EndIndent
	\end{algorithmic}
	\textbf{Output: }{$F_{MC}$} \Comment{MC filters necessary to maintain prediction of $I$ to inferred class $c$}
\end{algorithm}

\subsection{CFE model for MI filters}

To achieve our second objective of predicting MI filters, we follow a similar methodology used for MC filters but with key differences. The MI filters are those filters that, if they were altered to have higher magnitude, would have resulted in the model classifying the input to some other target class $c'_i \in C$, instead of the initially inferred class $c_i$ (or source class). \textcolor{black}{For this purpose, we train the MI CFE model to predict MI filters from $M$ that explain each image with respect to some target class $c'_i$.} 
The CFE model for MI filters is designed similarly to the CFE model for MC filters but with minor changes. The MI CFE model shares the feature extraction layers of $M$ up to the top convolution layer, as shown in Fig \ref{fig:block_daigram}.
\textcolor{black}{Given an input image $x_i$, model $M$ generates two outputs: (1) the feature maps $g_i \in [\mathbb{R}^+]^{1\times n}$ produced at the last convolution layer of $M$ after the global average pooling layer, and (2) the source or inferred class $c_i$. The MI CFE model takes as input feature maps $g_i$ and learns to generate non-binary MI filter map $F_{MI_i}$ corresponding to target class $c'_i$. This map is combined with the output of GAP layer of $M$ to modify the activation magnitudes with the objective to classify each input $x_i$ to the target class $c'_i$. With these definitions, the MI CFE model can be represented as:}
\begin{equation}\label{eq:new6}
\begin{aligned}
    \textcolor{black}{F_{MI_i}} & \textcolor{black}{= CFE_{MI}(g_i)} \\
    & \textcolor{black}{ = ReLU(d^n(g_i)),}
\end{aligned}
\end{equation}
\textcolor{black}{where $g_i$ represents the feature maps of $M(x_i)$ after global average pooling layer}, $d^n$ is a dense layer with $n$ units, and $ReLU$ is the ReLU activation function that produces the non-binary MI filter map $F_{MI_i}$. 

\textcolor{black}{The key difference between this equation and Eq. (\ref{eq:new4}) is the absence of sigmoid activation and the usage of standard ReLU instead of thresholded-ReLU.
Similar to the MC CFE model, the feature extraction layers shared between the MI CFE model and the pre-trained model $M$ are kept frozen during training of the CFE model. Only the dense layer $d^n$ weights are updated during training of the CFE model.} 
\textcolor{black}{To generate explanations, $CFE_{MI}$ model predicts the MI filters matrix for the corresponding input $x_i$.
$F_{MI_i}$ is added to $g$ to alter the filters from the top layer of $M$ after global average pooling. The DCNN model $M$ makes the alter prediction with the altered filters as $\hat{c_i}$:   
}
\begin{equation}\label{eq:new7}
    \textcolor{black}{\hat{c_i} = h( g_i + F_{MI_i}),}
\end{equation}
\textcolor{black}{where $h$ represents the classification (fully-connected and softmax) layers of $M$.}

The MI CFE model is trained to predict the optimal MI filter map $F_{MI_i}$ to reduce the loss between the predicted classes $\hat{c}_i$ and the target class $c'_i$. Optimal MI filters, $F_{MI_i}$, are learned by minimizing the following two losses: 1) cross-entropy (CE) loss, $L_{CE}$, for classifying each input image $x_i$ by modified model $M$ to the target class $c'_i$, and 2) sparsity loss, $L_{l1}$, that ensures $F_{MI_i}$ is sparse so that minimal filters are modified with minimal additive values such that the model $M$ classifies each input to class $c'_i$:
\begin{equation}\label{eq:5}
    L_{MI} = L_{CE}(\hat{c}_{i}, c'_i) + \lambda L_{l1}(F_{MI_i}, c'_i).
\end{equation}
This equation is similar to Eq. \ref{eq:1} with the difference that the logits loss is not used to find the MI filters. Logits loss, in this case, is not necessary as our objective can be achieved with just cross-entropy and sparsity losses. The first term of the loss function is the cross-entropy loss using which error between modified model $M$'s output and the target class $c'_i$ is minimized, which is computed using Eq. \ref{eq:2}. The second term minimizes the sum of the MI filter matrix $F_{MI_i}$ that pushes the CFE model to choose the least number of filters whose activation magnitude is increased minimally to predict each input $x_i$ to the target class $c'_i$. This loss can be computed using Eq. \ref{eq:3} by replacing $F_{MC_i}$ with $F_{MI_i}$.
\textcolor{black}{The procedure for MI CFE model training is summarized in Algorithm \ref{algo:2}}

\begin{algorithm}
	\caption{Procedure for training MI CFE model
	} 
	\label{algo:2}
	\textbf{Input: }{Image $I$, DCNN model $M$, target class $c'$, dataset $D$}
	\begin{algorithmic}[0]
	    \State 1. Train MI CFE model for target class $c'$ over the training dataset $D$
	    \Indent
	    \ForEach{image $x \in D$}
    	    \State \textcolor{black}{$g,c = M(x)$} 
    	    \State $\textcolor{black}{F_{MI} = ReLU(d^n(g))}$ 
    	    \State $\hat{c} = h( g + F_{MI})$ \Comment{alter prediction with just updated MI filters}
    	    \State{Minimize MI loss Equation \ref{eq:5}}
	        \Indent
    	        \State $L_{MI} = L_{CE}(\hat{c}, c') + \lambda L_{l1}(F_{MI}, c')$   
    	   \EndIndent
	    \EndFor
	    \EndIndent
	    \State 2. Generate counterfactual explanation using MI CFE model for input $I$
	    \Indent
    	    \State $g,c = M(I)$ 
    	    \State $F_{MI} = ReLU(d^n(g))$
        \EndIndent
	\end{algorithmic}
	\textbf{Output: }{$F_{MI}$} \Comment{MI addition to filters necessary to alter prediction of $I$ to target class $c'$}
\end{algorithm}

With both these CFE models for MC and MI filters, we can explain each decision of the pre-trained model $M$ in terms of finding the minimum required critical filters that maintain the model's decision to the inferred class or to finding the minimum set of filters that if they were altered with higher magnitude, would have classified the input to $c'_i$ instead of $c_i$. We show the importance of these filters by highlighting the features that activate them the most. The results of the proposed methodology are presented in the following section.

\section{Results}
\label{sec:four}

This section presents the results and discussion of the proposed counterfactual explanation (CFE) method. \textcolor{black}{We evaluate the explanations generated by the proposed CFE method qualitatively and quantitatively. In qualitative analysis, we provide visualization of the proposed method and show how to interpret these explanations. We compare these visualizations with existing counterfactual and contrastive explanation methods. Additionally, we conduct a user-study to evaluate the usefulness of the explanations provided based on the Explanation Satisfaction (ES) qualitative metric \cite{hoffman2018metrics}.} 

The results section is structured as follows.
In Section \ref{sub-1}, we discuss the experimental setup describing the dataset, the pre-trained model used for testing,
and the training details of the proposed CFE model for the explanation of the pre-trained model. \textcolor{black}{In Section \ref{sub-2}, we provide visualization of how the CFE method works and how it's output can be interpreted, and in Section \ref{sub-2-1}, we qualitatively compare the CFE method with \cite{selvaraju2017grad} and \cite{wang2020scout}. 
}
In Section \ref{sub-3}, we present the quantitative evaluation of the proposed method \textcolor{black}{where we measure the impact of disabling the MC filters for different classes on the overall model accuracy and class recall. We also include an analysis on using different weights for the sparsity loss and measure the effectiveness of logits loss. Finally, in Section \ref{sub-compSOTA}, we compare our method with state-of-the-art explanation methods.}

\subsection{Experimental setup} \label{sub-1}

For the evaluation of the proposed CFE method, we used the Caltech-UCSD Birds (CUB) 2011 \cite{WahCUB_200_2011} dataset. We train a VGG-16 \cite{simonyan2014very} model on this dataset and train our CFE model to provide explanations for the trained model's decisions. 
The VGG-16 model was initially trained by removing the dense classification layers and adding a GAP layer after the top convolution layer, followed by a dropout and the output softmax layer, as discussed in Section \ref{sec:three}.  
The VGG-16 model was trained in two steps. First, we performed transfer learning to train the newly added output softmax layer with stochastic gradient descent (SGD) optimizer using imageNet \cite{ILSVRC15} pre-trained weights. Transfer learning was performed for 50 epochs with a 0.001 learning rate, 0.9 momentum, 32 batch size, and 50\% dropout without data augmentation. \textcolor{black}{In the second step, we fine-tuned all model layers for 150 epochs at a 0.0001 learning rate with standard data augmentation and kept all other parameters the same.} The VGG-16 model achieved the final training and testing accuracy of 99.0\% and 69.5\%, respectively. 

\subsubsection{Counterfactual explanation model training details}

The CFE model provides contrastive and counterfactual explanations of the decisions of the pre-trained VGG-16 model by predicting the minimum correct (MC) and minimum incorrect (MI) filters for each decision by the model with respect to the original inferred class (source class) and some target alter class.
The CFE models for MC and MI filter prediction are comprised of similar architectures as the pre-trained model but with some differences, as discussed in Section \ref{sec:three}. For the MC CFE model, the feature extraction layers are frozen. The output layer consists of the same number of units as the number of filters in the top convolution layer, with a sigmoid activation function followed by thresholded-ReLU. The weights of the feature extraction layers are shared with the pre-trained model being explained. The CFE model is trained for a given alter class by minimizing the three losses, namely, cross-entropy loss, sparsity loss, and logits loss, as discussed in Section \ref{sec:three}.
The MI CFE model follows similar architecture but with the difference that the output dense layer after the top convolution layer is followed by ReLU activation function instead of sigmoid activation, and there is no thresholded-ReLU layer. The MI CFE model is trained for a target class by minimizing the two losses, i.e., cross-entropy loss and sparsity loss, as discussed in Section \ref{sec:three}.
The MC and MI CFE models are trained using SGD optimizer with 0.001 learning rate, 0.9 momentum, and 32 batch size for 200 epochs. 
The weight for sparsity loss is chosen as $\lambda = 2$ for MC CFE models, and $\lambda = 1$ for MI CFE model. 

\subsection{\textcolor{black}{Qualitative analysis}} \label{sub-2}

\begin{figure*}
\begin{center}

\includegraphics[width=1.0\linewidth]{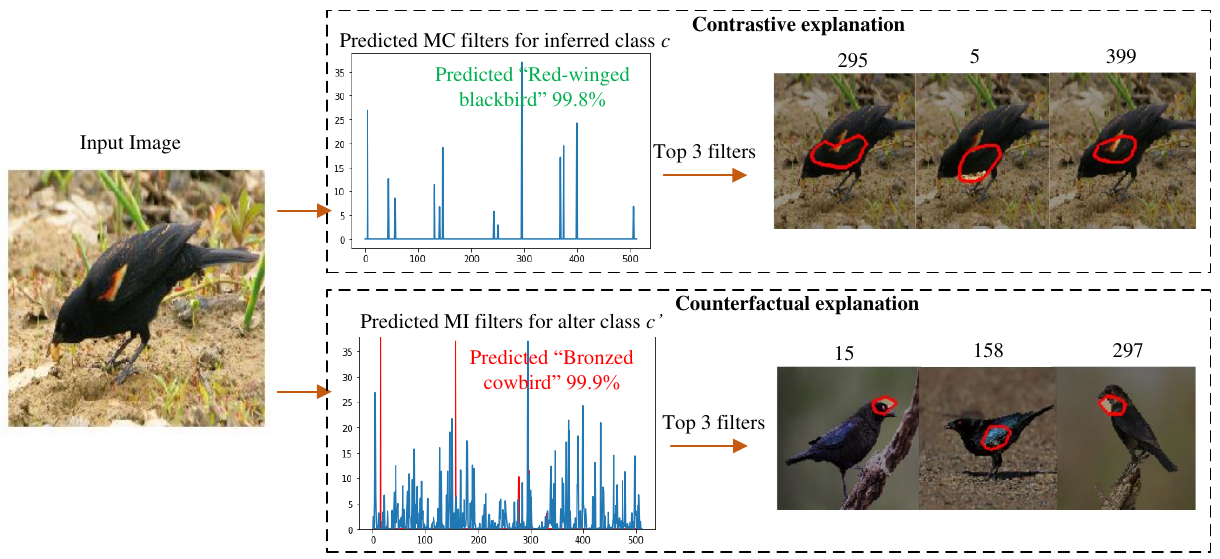}

    \caption{\textcolor{black}{Contrastive and counterfactual explanations for a sample image from CUB dataset. Contrastive explanation highlights the top-3 MC filters representing features important for the inferred class. Counterfactual explanation highlights top-3 MI filters that represent features required for the alter class.}}
    
    \label{fig:results_1}

\end{center}

\end{figure*}

\textcolor{black}{In this section, we qualitatively discuss the results of the proposed CFE approach by providing the visualization of the MC and MI filters, comparing explanations with existing methods, performing misclassification analysis, and conducting a user evaluation to assess the usefulness of different explanation methods. }

\subsubsection{\textcolor{black}{Explanation visualization and interpretation}}

Fig. \ref{fig:results_1} shows the CFE model results for a sample image from the CUB dataset that was correctly predicted as ``Red-winged blackbird" with 99.9\% probability by the VGG-16 model. 
Our model highlights the MC and MI filters necessary for classifying the input image either to the source class or to the target class, respectively.
\textcolor{black}{The contrastive explanation in Fig. \ref{fig:results_1}, identifies the most important MC filters plotted as a graph of their activation magnitudes (y-axis) against filter number (x-axis). Using these filters, the model maintains the prediction of the input image to its inferred class, i.e. ``Red-winged Blackbird" with 99.8\% probability}. 

We visualize the concepts represented by the top-3 filters (based on activation magnitude) by drawing the filter's receptive field (RF) \cite{zhou2014object} on the input image. RF is the image-resolution feature map of the selected filter from the top convolution layer that allows us to understand where the filter pays most attention. 
\textcolor{black}{RF for filters 295 and 399 in Fig. \ref{fig:results_1} show that they focus around the ``red spot" of the bird, while filter 5 focuses on the belly of the bird. 
The counterfactual explanation in Fig. \ref{fig:results_1} identifies the MI filters with respect to the ``Bronzed cowbird" class (top-3 predicted class) shown as a graph of filter activation magnitudes against filter number. The MI filters are highlighted in red. The CFE model predicts the minimum additive values to modify these MI filters such that it results in the input image being classified to the target class, i.e., ``Bronzed cowbird".} \textcolor{black}{ The RF visualization of the top-3 MI filters show the most important features for the target class.} Filter 15 activates the most on the red-colored eye of the ``Bronzed cowbird". In contrast, filter 158 activates the most on the blueish tinge on the bird's wings. These two filters identify the critical features that if they were present in the input image, the model would have been more likely to classify the bird as ``Bronzed cowbird" instead of the ``Red-winged blackbird." Filter 297, on the other hand, activates the most on the bird's black neck. However, it can be observed that this feature is common to both birds. Therefore it is not enough to discriminate between the two classes, and it is not activated with higher magnitude.
\begin{figure}[hbt!]
    \centering

    \subfloat[] {\includegraphics[width=.250\textwidth]{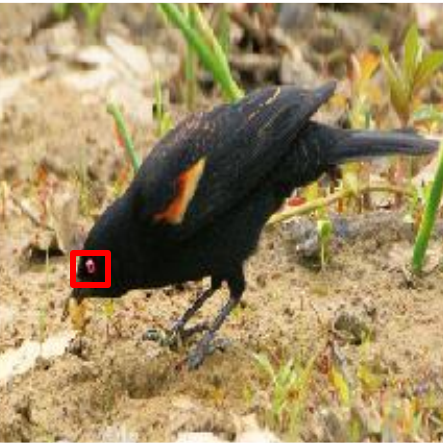} \label{fig:results_2-a}}
    \subfloat[] {\includegraphics[width=.250\textwidth]{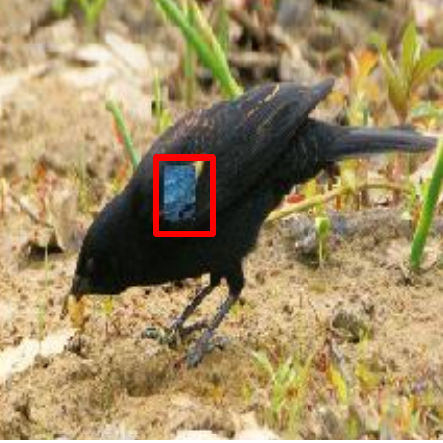} \label{fig:results_2-b}}
    \subfloat[] {\includegraphics[width=.250\textwidth]{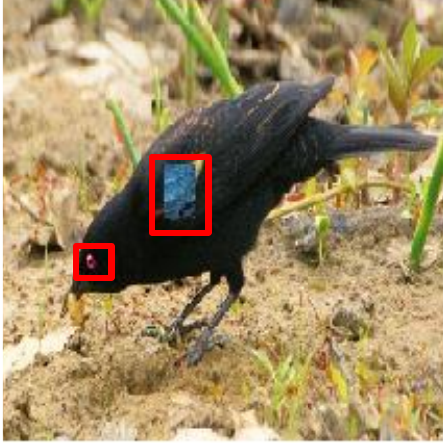} \label{fig:results_2-c}}
        

    \caption{Artificially introducing the most important features relevant to the alter class identified in Fig. \ref{fig:results_1} (Bronzed cowbird). (a) Adding eye color. (b) Adding wing pattern. (c) Adding both eye color and wing pattern.} 
        \label{fig:results_2}
\end{figure}

\textcolor{black}{To show that, in case the features represented by filters 15 and 158 in Fig. \ref{fig:results_1}  were present in the input image, the model would have been more likely to classify the image as ``Bronzed cowbird"}, we modify the input image to artificially introduce these features, as shown in Fig. \ref{fig:results_2}. In Fig. \ref{fig:results_2-a}, artificially changing the color of the eye to match the other bird's eye color alone was not enough to change the model prediction. However, if we introduce a patch of bluish tinge to the bird's wing, as shown in Fig. \ref{fig:results_2-b}, the model changes its prediction to ``Bronzed cowbird" with a probability of 53\%. In Fig. \ref{fig:results_2-c}, we modified both the eye color and added bluish tinge on the wing of the bird that resulted in the model classifying the image as ``Bronzed cowbird" instead of the ``Red-winged blackbird" with 82\% confidence, thus highlighting the importance of the MI filters predicted by the CFE model for the target class. We can say that these features separate the model's decision between the two classes.

\begin{figure*}[t!]
\begin{center}
\includegraphics[width=.7\textwidth]{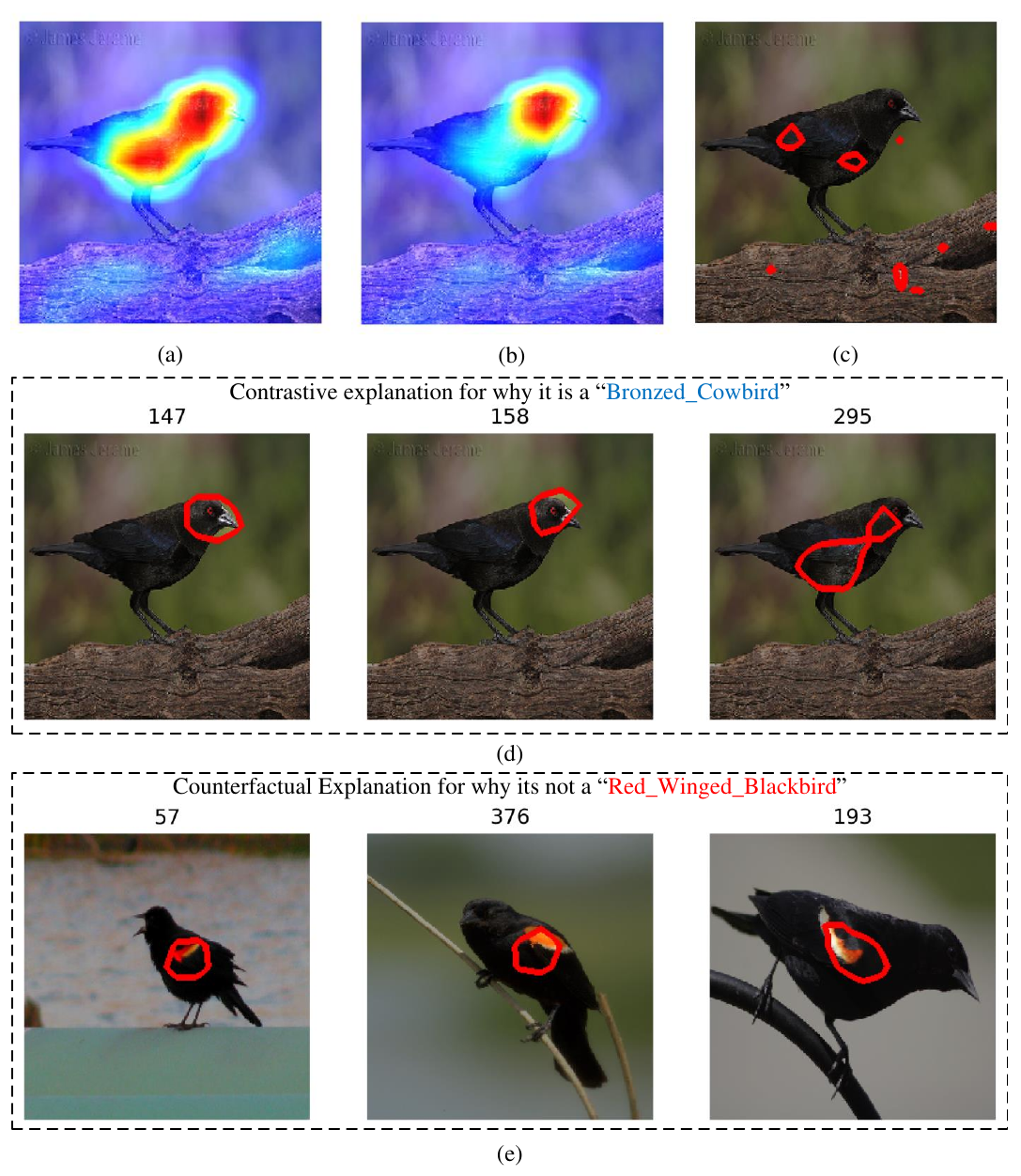}

    \caption{\textcolor{black}{Explanation comparison of GradCAM \cite{selvaraju2017grad}, SCOUT \cite{wang2020scout}, and proposed CFE method for a query image that was classified as ``Bronzed cowbird", while the alter class is set as ``Red-winged blackbird". (a) GradCAM explanation for the inferred class (Bronzed cowbird). (b) GradCAM explanation for the alter class. (c) SCOUT explanation for why the image is classified to inferred class and not to alter class. (d) Contrastive explanation of the proposed CFE method. (e) Counterfactual explanation of the proposed CFE method.}}
    
    \label{fig:visual_comparison}
\end{center}
\end{figure*}

\textcolor{black}{To further illustrate the effectiveness of the proposed method, a visual comparison of explanations provided by GradCAM \cite{selvaraju2017grad}, SCOUT \cite{wang2020scout}, and the proposed CFE method is shown in Fig. \ref{fig:visual_comparison}. Fig. \ref{fig:visual_comparison} shows explanations for a test image of the ``Bronzed cowbird" class that the DCNN correctly classified. Fig. \ref{fig:visual_comparison}a shows GradCAM explanation for the inferred class by highlighting the head, neck, and belly regions as the important features for that class. However, in Fig. \ref{fig:visual_comparison}b, GradCAM fails to provide useful information regarding the alter class since it only highlights the head part of the bird, which is a similar feature between the two classes. In Fig. \ref{fig:visual_comparison}c, SCOUT fails to show a meaningful explanation for why the bird is a ``Bronzed cowbird" and not a ``Red-winged blackbird" since it highlights indistinguishable features. \textcolor{black}{Another drawback of methods like SCOUT and GradCAM is that they provide explanations for different inferred and target classes based on the same input image. If we want to analyze why a model classified an image to class A and not to class B, we must switch the target class for the same input image to see where the model is looking for both classes on the same input. This limits the capability of methods like GradCAM when the input image has only one object and produces less meaningful explanation. Our proposed CFE model, on the other hand, does not have this limitation. The proposed CFE method identifies filters relevant to the inferred class as well as filters that if they were active, the model would have been likely to classify the input to another target class. We can visualize what features the filters represent by looking at their receptive fields on any random image from the inferred or target class. In this regard, the Figs. \ref{fig:visual_comparison}d and \ref{fig:visual_comparison}e, of the proposed CFE method show clear explanations.} The contrastive explanation in Fig. \ref{fig:visual_comparison}d shows that the bird's head and beak along with the eye are particularly discriminating features for it, highlighted by filters 147 and 158, respectively. The counterfactual explanation in Fig. \ref{fig:visual_comparison}e shows that the red spot on the wings is the main feature absent from the ``Bronzed cowbird" that would have been important for the image to be classified to the alter class. Together, both the contrastive and counterfactual explanations effectively highlight the DCNN model's reasoning for the classification example.} \textcolor{black}{Additional visualizations and comparisons are provided as supplementary material.}

\begin{figure}[hbt!]
    \centering
    
    \subfloat[] {\includegraphics[width=.70\textwidth]{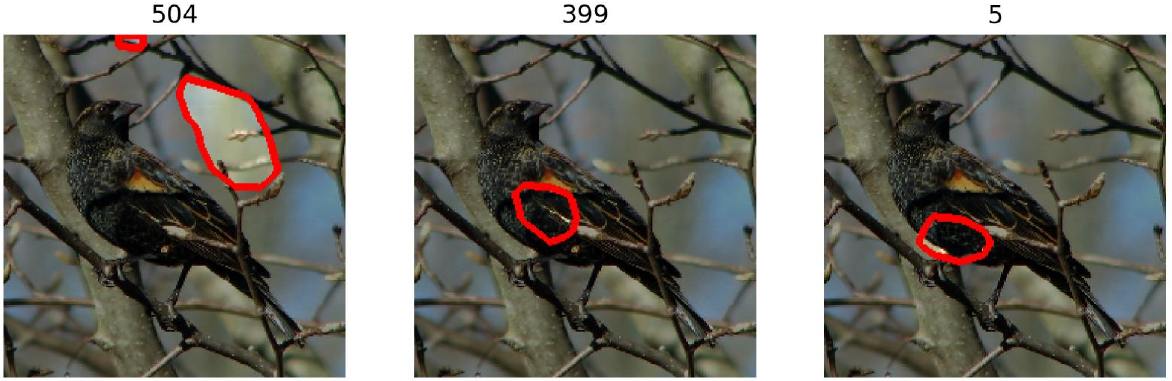} \label{fig:results_3-a}}\hfill
    \subfloat[] {\includegraphics[width=.70\textwidth]{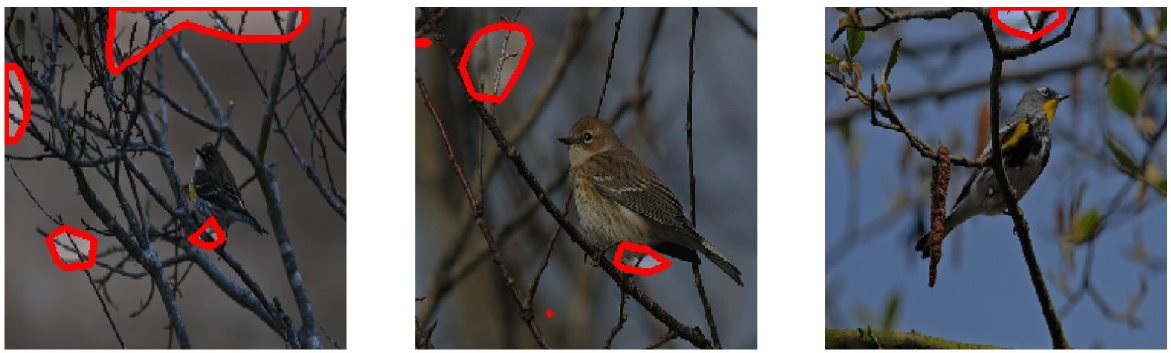} \label{fig:results_3-b}}\hfill
    \subfloat[] {\includegraphics[width=.70\textwidth]{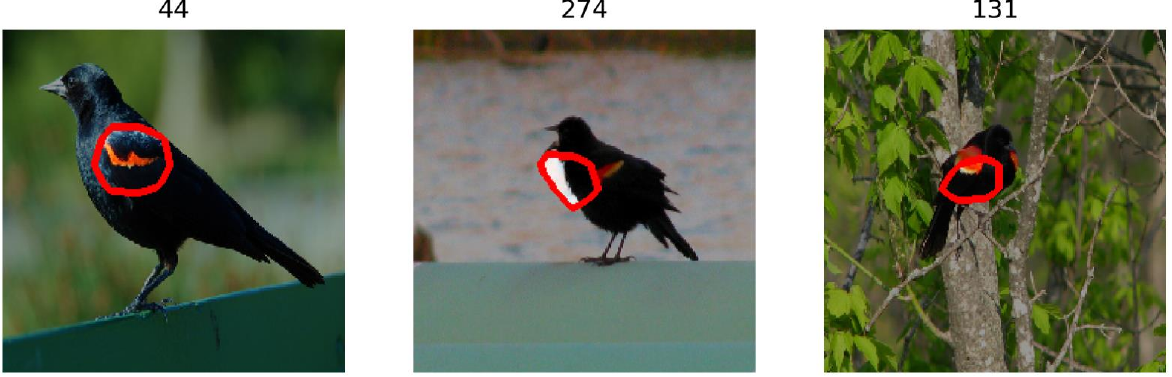} \label{fig:results_3-c}}
        

    \caption{\textcolor{black}{Identifying the erroneous filters that result in misclassification. (a) Input image of class ``Red-winged blackbird" with RFs of top-3 MC filters involved in incorrect classification of the image as ``Myrtle warbler". (b) Top-3 images from the inferred class that activate filter 504 the most. (c) Top-3 MI filters with RF visualization on images from the true class that activate them the most.}} 
        \label{fig:results_3}

\end{figure}

\begin{figure}[]
    \centering
    \subfloat[] {\includegraphics[width=.35\textwidth]{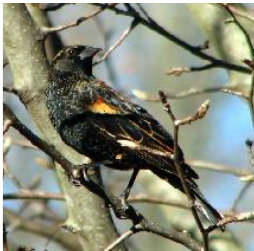} \label{fig:results_3_1-a}}
    \subfloat[] {\includegraphics[width=.35\textwidth]{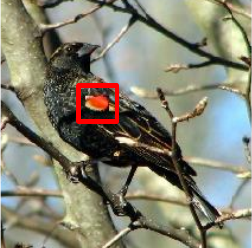} \label{fig:results_3_1-b}}


    \caption{Modifying the misclassified input image according to features identified by MI filters. (a) Image of class ``Red-winged blackbird" misclassified as ``Myrtle warbler" with 92\% probability. (b) Modified image that is correctly classified as ``Red-winged blackbird" with 96\% probability.}
        \label{fig:results_3_1}
\end{figure}

\subsubsection{Misclassification analysis}

One of the useful applications of the proposed explanation method is misclassification analysis. Since our model predicts the MC and MI filters for an input sample with respect to different classes, it is possible to analyze why the model has classified an input to the inferred class and why not to some other, maybe top-2 or 3 class. In Fig. \ref{fig:results_3}, we show a misclassification case where an image of class ``Red-winged blackbird" was incorrectly classified as ``Myrtle warbler" with a probability of 92\%. \textcolor{black}{In Fig\ref{fig:results_3-a}, we show the top-3 MC filters for the inferred class with RF drawn on the input image for visualization. As it turns out, the most highly activated filter (filter 504) focuses on the background region around the tree's branches. In Fig. \ref{fig:results_3-b}, we show that filter 504 is associated with the incorrectly inferred class by drawing RF on the top-3 images from the inferred class that activate it the most.} It can be seen that this filter mostly activates on the background branches making it an unreliable filter, and this decision can be treated as an untrustworthy or a misclassification case.

On the other hand, the CFE model predicted 13 MI filters from the original filter activations with respect to the actual class, using which the model correctly predicted the image as ``Red-winged blackbird" with a probability of 91\%. \textcolor{black}{RF visualization of the top-3 MI filters out of the 13 are shown in Fig. \ref{fig:results_3-c}, with RF drawn on images from the true class that highly activate them.} Filters 44 and 131 in Fig. \ref{fig:results_3-c} suggest that the input image should have a red-colored spot on the wing of the bird to classify it correctly to the true class. However, the wing of this particular bird has an orange-colored spot instead of red color. If we manually change the color of this spot to red or replace it with a similar-sized red spot from another bird of this class, then the model can correctly classify it to the actual class, as shown in Fig. \ref{fig:results_3_1-b}. Introducing the red spot on the bird's wing changed the model's prediction to true class with 96\% probability.

In summary, we can say that the original image in Fig. \ref{fig:results_3_1-a} was incorrectly classified as ``Myrtle warbler" because the model paid more attention to the tree branches in the background that were common in images with ``Myrtle warbler" bird. Moreover, the bird does not have a proper red color spot that is important for being classified as a ``Red-winged blackbird". These two issues contributed mainly to the incorrect classification. If the image had a proper red-colored spot, which is the defining characteristic of the true class of this image, or if the background did not have such branches, then the model would have been more likely to classify the image correctly.
\textcolor{black}{Such types of explanations provided by our model help to uncover the decision process behind the pre-trained black-box models, making them more transparent and improving trust in their use.}

\begin{figure}[t!]
\begin{center}
\includegraphics[width=.7\textwidth]{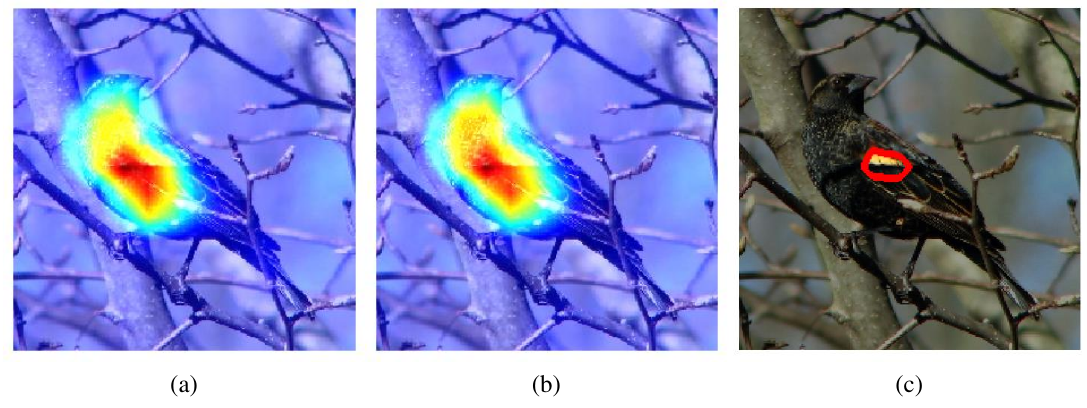}

    \caption{\textcolor{black}{Misclassification case explanations of GradCAM \cite{selvaraju2017grad} and SCOUT \cite{wang2020scout}. (a) GradCAM explanation for the incorrect inferred class (Myrtle warbler). (b) GradCAM explanation for the true class (Red-winged blackbird). (c) SCOUT explanation for why the image is classified to inferred class and not to true class. }}
    
    \label{fig:misclassification_comparison}
\end{center}
\end{figure}

\textcolor{black}{In contrast, the explanations provided by GradCAM \cite{selvaraju2017grad} and SCOUT \cite{wang2020scout} for this misclassification case are shown in Fig. \ref{fig:misclassification_comparison}. Figs. \ref{fig:misclassification_comparison}a and \ref{fig:misclassification_comparison}b show the GradCAM explanations for the incorrect inferred class (Myrtle warbler) and the true class (Red-winged blackbird), respectively. It can be observed that these explanations highlight similar regions as evidence for the target class in each case which fails to provide meaningful reasons for the model's decision. In Fig. \ref{fig:misclassification_comparison}c, the SCOUT explanation highlights the red/orange spot as the region that separates the model's decision in classifying the input as ``Myrtle warbler" and not ``Red-winged blackbird." This region, however, is not strong evidence for the ``Myrtle warbler" class since it is important for the true class, ``Red-winged blackbird." This explanation fails to provide reasons for why the model incorrectly classified the image.}

\subsection{\textcolor{black}{User evaluation}} \label{sub-2-1}
\textcolor{black}{To show the effectiveness of the explanations provided by the proposed CFE method, we conducted a user-study to qualitatively evaluate different counterfactual and contrastive explanation methods including \cite{selvaraju2017grad} and \cite{wang2020scout} in terms of the Explanation Satisfaction (ES) \cite{hoffman2018metrics} qualitative metric. This metric was previously used by \cite{akula2020cocox} to perform user-study, and we have followed their protocol closely in this work as well. ES metric measures the user's satisfaction at achieving an understanding of the DCNN model based on different explanations provided in terms of metrics such as usefulness, understandability, and confidence \cite{hoffman2018metrics}.} 

\textcolor{black}{The user-study is conducted by creating two expert and non-expert human subjects groups. The non-expert group consists of 30 subjects with a limited understanding of the computer vision field, whereas the expert group consists of 10 subjects that routinely train and evaluate DCNN models. Subjects in each group go through a familiarization phase where they are first shown a sample query image and the DCNN model's classification decision for that image. The users are then shown various images from the predicted class, and some alter class selected from top-2 or top-3 classes to help them understand the differences between the two classes. The users are then shown explanations generated by \cite{selvaraju2017grad}, \cite{wang2020scout}, and the proposed method for why the model classified the image to the inferred class and not to the alter class, along with a brief description on how to interpret the generated explanations.
Finally, the users are shown the DCNN model's decisions on 10 test images, along with the explanations for these provided by the explanation techniques regarding the inferred and alter classes. At the end of the test examples, the users are asked to rate each explanation method in terms of ES metrics of understandability, usefulness, and confidence on a Likert scale of 0 to 4. Table \ref{tab:user-evaluation} presents the results. }

\textcolor{black}{The findings in Table \ref{tab:user-evaluation} highlight that the users from both the expert and non-expert groups found the explanations provided by the proposed method to be beneficial and understandable. Particularly for the expert group, there is a significant difference in the level of satisfaction achieved compared to the non-expert group. The reason for this can be because the proposed method probes the internal working of the DCNN to provide an explanation in the form of filters and their visualizations, which is a concept easily understandable by expert users. 
For non-expert users, the explanations provided by the GradCAM method may appear to be sufficient since they can be visually pleasing, and due to this reason, there is less gap between the understandability scores of GradCAM and the proposed method for non-expert users.}

\textcolor{black}{Furthermore, as evident from high scores given to our approach by expert users, our approach caters to the need of DCNN model developers who already have some understanding of how DCNNs work. For these users, our approach identifies what filters are critical for the model’s decision-making process and shows what concepts and features these filters represent. These users can make informed decisions regarding the model’s weaknesses and trustworthiness and judge its overall performance in real-world scenarios. Thus, leading to the success of the proposed explanation method in terms of understandability, usefulness, and confidence.}

\begin{table}[]
\footnotesize
\centering
\caption{\textcolor{black}{Qualitative evaluation based on Explanation Satisfaction metric for GradCAM, SCOUT, and the proposed CFE methods.}}
\label{tab:user-evaluation}
\begin{tabular}{|c|ccc|}
\hline
\multirow{3}{*}{\textbf{Explanation}} & \multicolumn{3}{c|}{\textbf{Explanation Satisfaction ($\pm$std)}} \\ \cline{2-4} 
 & \multicolumn{1}{c|}{Understandability} & \multicolumn{1}{c|}{Usefulness} & Confidence \\ \cline{2-4} 
 & \multicolumn{3}{c|}{\textbf{Non-expert users}} \\ \hline
Grad-CAM \cite{selvaraju2017grad} & \multicolumn{1}{c|}{3.7 ($\pm$1.1)} & \multicolumn{1}{c|}{3.6 ($\pm$1.1)} & 3.6 ($\pm$1.1) \\ \hline
SCOUT \cite{wang2020scout} & \multicolumn{1}{c|}{3.4 ($\pm$1.0)} & \multicolumn{1}{c|}{3.1 ($\pm$1.1)} & 3.3 ($\pm$1.0) \\ \hline
CFE (proposed) & \multicolumn{1}{c|}{\textbf{3.9 ($\pm$0.9)}} & \multicolumn{1}{c|}{\textbf{3.8 ($\pm$0.9)}} & \textbf{4.0 ($\pm$0.9)} \\ \hline
 & \multicolumn{3}{c|}{\textbf{Expert users}} \\ \hline
Grad-CAM \cite{selvaraju2017grad} & \multicolumn{1}{c|}{3.7 ($\pm$1.2)} & \multicolumn{1}{c|}{3.5 ($\pm$1.1)} & 3.6 ($\pm$1.2) \\ \hline
SCOUT \cite{wang2020scout} & \multicolumn{1}{c|}{3.0 ($\pm$1.4)} & \multicolumn{1}{c|}{2.8 ($\pm$1.5)} & 2.8 ($\pm$1.4) \\ \hline
CFE (proposed) & \multicolumn{1}{c|}{\textbf{4.5 ($\pm$0.6)}} & \multicolumn{1}{c|}{\textbf{4.2 ($\pm$0.8)}} & \textbf{4.4 ($\pm$0.6)} \\ \hline
\end{tabular}
\end{table}

\subsection{Quantitative analysis} \label{sub-3}

In this section, we quantitatively discuss the results of the proposed CFE methodology in terms of finding the most commonly activated MC filters predicted by the CFE model for explaining the pre-trained VGG-16 model with respect to different classes. We show the importance of these MC filters by demonstrating the effect of disabling them on the class recall metric compared to the effect on the overall model accuracy. Furthermore, we also discuss the effect of different training parameters on the MC filters predicted and CFE model accuracy for explaining the pre-trained VGG-16 model.

\subsubsection{Activated filter statistics}

\begin{figure*}[hbt!]
\centering

    \subfloat[] {\includegraphics[width=.50\textwidth]{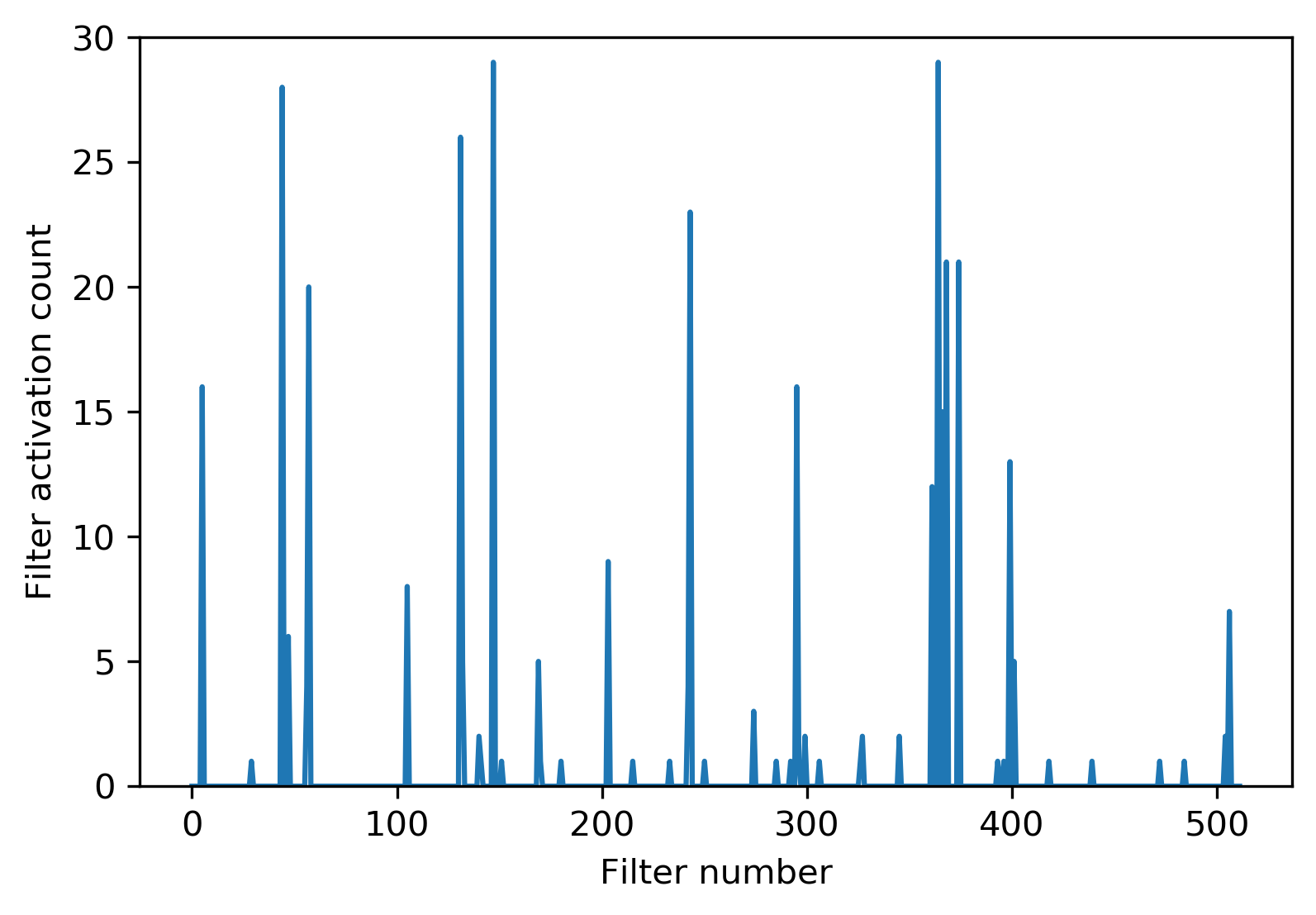}\label{fig:stats_vgg_cub-c}}
    \subfloat[] {\includegraphics[width=.50\textwidth]{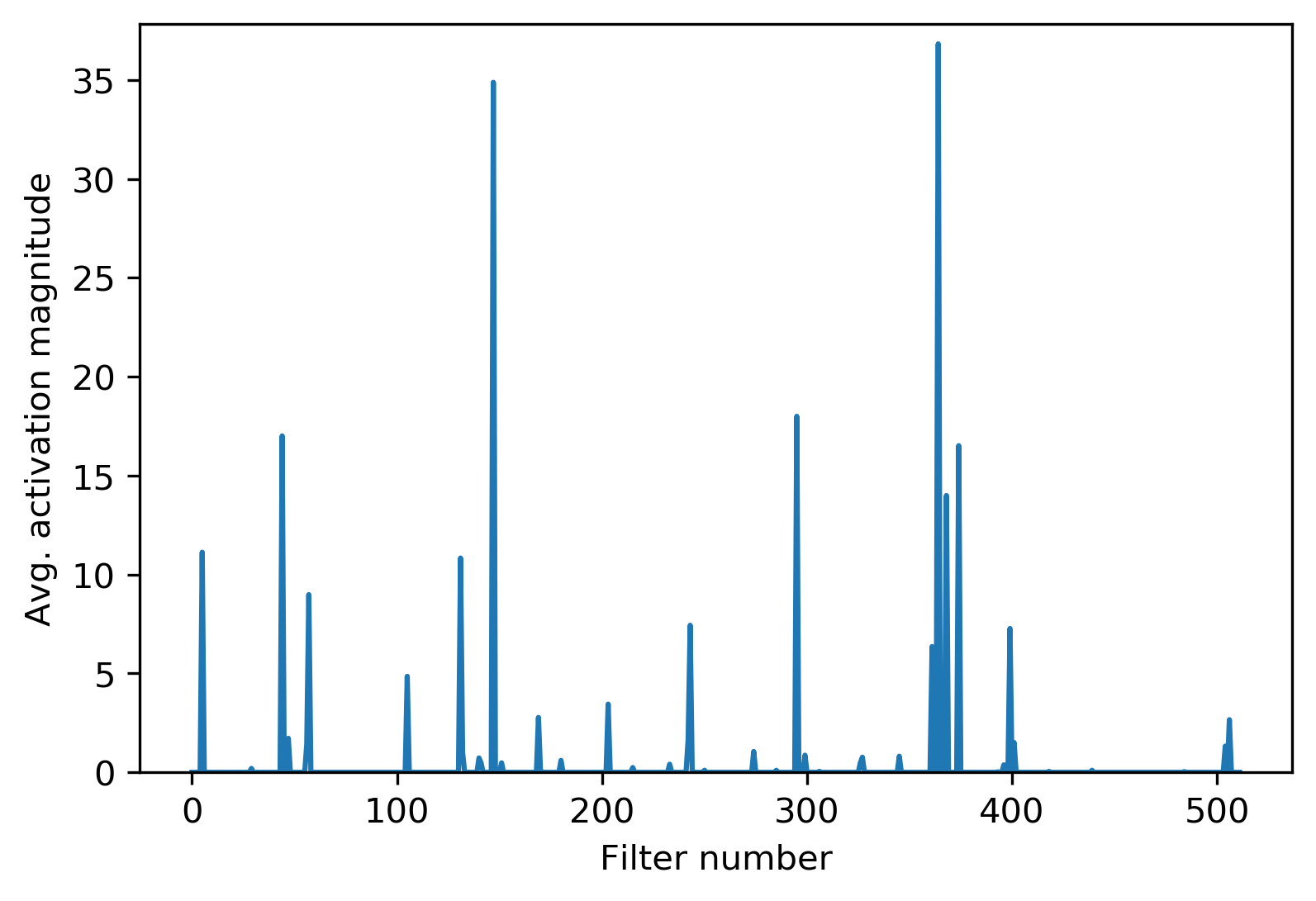}\label{fig:stats_vgg_cub-d}}


    \caption{Statistical analysis of MC filters predicted by CFE model for explaining VGG-16 model with respect to ``Red-winged blackbird' class. (a) Filter activation count of MC filters for all test images of class ``Red-winged blackbird'. X-axis represents filter numbers and y-axis is the filter activation count. (b) Normalized filter activation magnitude for the MC filters.}
        \label{fig:stats_vgg_cub}
\end{figure*}

First, we discuss the filter activation statistics of the MC filters predicted by the CFE model for explaining the pre-trained VGG-16 model with respect to different classes. For a given class, we use the CFE model to predict MC filters for all test images of that class. We then accumulate the predicted MC filters to find the number of times each filter is predicted and compute the normalized activation magnitude for those filters.
Fig. \ref{fig:stats_vgg_cub} shows the activated filter analysis of MC filters for ``Red-winged blackbird" class. There are 5794 test images in the CUB dataset, of which 30 belong to this class. 
Fig. \ref{fig:stats_vgg_cub-c} shows that filters 44, 147, and 364 are predicted as part of MC filters for nearly all of the test images. Fig. \ref{fig:stats_vgg_cub-d} shows the normalized activation magnitudes of these filters. It can be seen that filters 147 and 364 have, on average, the highest activation magnitude of these filters. In other words, these filters are globally the most important filters for this class and represent crucial features/concepts relevant to this class.  
\textcolor{black}{We show the importance of these globally significant filters for the ``Red-winged blackbird" class by disabling them from the pre-trained VGG-16 model and reporting the decrease in the model's ability to accurately classify images of this class as compared to the overall accuracy of the model. Interestingly, disabling all 31 MC filters predicted by the CFE model for the ``Red-winged blackbird" class reduced the class recall from 93.3\% to 30\%. In contrast, the overall model accuracy is decreased by less than 2\%. Similarly, we carry out this analysis for a few other classes and summarize the findings in Table \ref{tab:filter_analysis}. In these cases, \textcolor{black}{disabling around 31-44} globally critical MC filters results in a significant decrease in the class recall, whereas the overall model accuracy is reduced by just 2-3\%. On the contrary, it can be seen that randomly disabling 40 filters has a negligible effect on class recall. This analysis shows that the MC filters predicted by the CFE model for a particular class represent features exclusive to that class, and disabling them affects the overall model accuracy minimally while significantly reducing the class recall score.
}

\begin{table}[]
\footnotesize
\centering
\caption{Affect of disabling global MC filters on class recall metric for different classes}
\label{tab:filter_analysis}
\begin{tabular}{|l|l|l|l|l|l|}
\hline
\multirow{2}{*}{\textbf{Class}} & \multirow{2}{*}{\textbf{\begin{tabular}[c]{@{}l@{}}No. of Global \\ MC filters \\ disabled\end{tabular}}} & \multicolumn{3}{c|}{\textbf{Class recall (\%)}} & \multirow{2}{*}{\textbf{\begin{tabular}[c]{@{}l@{}}Model acc.\\ (Orig. = \\ 69.5\%)\end{tabular}}} \\ \cline{3-5}
 &  & \textbf{Orig.} & \textbf{\begin{tabular}[c]{@{}l@{}}Rand. \\ disabled\end{tabular}} & \textbf{\begin{tabular}[c]{@{}l@{}}MC \\ disabled\end{tabular}} &  \\ \hline
\begin{tabular}[c]{@{}l@{}}Red-winged \\ blackbird\end{tabular} & 31 & 93.3 & 93.3 & 30.0 & 67.6 \\ \hline
Bronzed cowbird & 43 & 73.3 & 73.3 & 3.3 & 67.4 \\ \hline
American redstart & 33 & 80.0 & 76.7 & 23.3 & 66.0 \\ \hline
\begin{tabular}[c]{@{}l@{}}Nelson sharp-\\ tailed sparrow\end{tabular} & 44 & 53.3 & 56.7 & 0.0 & 67.7 \\ \hline
Myrtle warbler & 37 & 53.3 & 40.0 & 6.7 & 67.7 \\ \hline
\end{tabular}
\end{table}

\subsubsection{Trade-off between CFE model accuracy and predicted filter sparsity}
In this section, we will discuss the effect on CFE model training \textcolor{black}{and testing} accuracy by varying the losses and training parameters and also report the average number of MC filters predicted by the CFE model with these changes. Table \ref{tab:L1_loss} presents the effect of different weights assigned to the sparsity loss in Eq. \ref{eq:1} for explaining the VGG-16 model with respect to the ``Red-winged Blackbird" class using MC filters. 
It can be seen that as the sparsity loss weight is increased from $\lambda = 1 $ to $ \lambda = 4$, the average number of filters predicted for classifying all \textcolor{black}{train or test images in the CUB dataset to the ``Red-winged Blackbird" decreases from 20.5 and 21.9 to 10.1 and 11.4, respectively. Although fewer predicted filters are desirable, it comes at the cost of higher cross-entropy loss that results in low confidence predictions.} 

Table \ref{tab:logits_loss} presents a similar analysis where we show the effect of training the MC CFE model with and without the logits loss. Without using logits loss, the CFE model suffered higher training \textcolor{black}{and testing loss but better accuracies}. This means that the sparse filters predicted by the CFE model had a lower impact or were less critical towards the specified class, leading to low confidence predictions. With logits loss, on the other hand, the CFE model predicted \textcolor{black}{on average 0.8 and 1.1 more filters, respectively, for training and testing sets, resulting in lower CE loss but slightly decreased accuracy.} However, the predicted filters are more likely to be relevant to the specified alter class.

\begin{table}[]
\footnotesize
\centering
\textcolor{black}{
\caption{\textcolor{black}{Sparsity loss analysis for MC CFE model for ``Red-winged blackbird" class}}
\label{tab:L1_loss}
\begin{tabular}{|c|lllll|}
\hline
\multicolumn{1}{|l|}{\textbf{CFE model}} & \multicolumn{1}{l|}{\textbf{$\lambda$}} & \multicolumn{1}{l|}{\textbf{Accuracy}} & \multicolumn{1}{l|}{\textbf{CE loss}} & \multicolumn{1}{l|}{\textbf{L1 loss}} & \textbf{Filters} \\ \hline
\multirow{8}{*}{VGG-16} & \multicolumn{5}{c|}{\textbf{Training}} \\ \cline{2-6} 
 & \multicolumn{1}{l|}{1} & \multicolumn{1}{l|}{99.8\%} & \multicolumn{1}{l|}{0.246} & \multicolumn{1}{l|}{0.594} & 20.5 \\ \cline{2-6} 
 & \multicolumn{1}{l|}{2} & \multicolumn{1}{l|}{99.4\%} & \multicolumn{1}{l|}{0.439} & \multicolumn{1}{l|}{0.421} & 14.5 \\ \cline{2-6} 
 & \multicolumn{1}{l|}{4} & \multicolumn{1}{l|}{98.5\%} & \multicolumn{1}{l|}{0.839} & \multicolumn{1}{l|}{0.293} & 10.1 \\ \cline{2-6} 
 & \multicolumn{5}{c|}{\textbf{Testing}} \\ \cline{2-6} 
 & \multicolumn{1}{l|}{1} & \multicolumn{1}{l|}{99.3\%} & \multicolumn{1}{l|}{0.341} & \multicolumn{1}{l|}{0.618} & 21.9 \\ \cline{2-6} 
 & \multicolumn{1}{l|}{2} & \multicolumn{1}{l|}{99.1\%} & \multicolumn{1}{l|}{0.568} & \multicolumn{1}{l|}{0.449} & 16.0 \\ \cline{2-6} 
 & \multicolumn{1}{l|}{4} & \multicolumn{1}{l|}{98.1\%} & \multicolumn{1}{l|}{0.973} & \multicolumn{1}{l|}{0.317} & 11.4 \\ \hline
\end{tabular}
}
\end{table}

\begin{table}[]
\footnotesize
\centering
\caption{\textcolor{black}{MC CFE model training analysis with and without logits loss using fixed $\lambda = 2$ for ``Red-winged blackbird" class}}
\textcolor{black}{
\label{tab:logits_loss}
\begin{tabular}{|l|lllll|}
\hline
\textbf{VGG-16 CFE} & \multicolumn{1}{l|}{\textbf{Accuracy}} & \multicolumn{1}{l|}{\textbf{CE Loss}} & \multicolumn{1}{l|}{\textbf{L1 loss}} & \multicolumn{1}{l|}{\textbf{Logits loss}} & \textbf{Filters} \\ \hline
 & \multicolumn{5}{c|}{\textbf{Training}} \\ \hline
With Logits loss & \multicolumn{1}{l|}{99.4\%} & \multicolumn{1}{l|}{0.439} & \multicolumn{1}{l|}{0.421} & \multicolumn{1}{l|}{-0.313} & 14.5 \\ \hline
W/o logits loss & \multicolumn{1}{l|}{99.5\%} & \multicolumn{1}{l|}{0.529} & \multicolumn{1}{l|}{0.395} & \multicolumn{1}{l|}{-0.275} & 13.7 \\ \hline
 & \multicolumn{5}{c|}{\textbf{Testing}} \\ \hline
With Logits loss & \multicolumn{1}{l|}{99.1\%} & \multicolumn{1}{l|}{0.568} & \multicolumn{1}{l|}{0.449} & \multicolumn{1}{l|}{-0.299} & 16.0 \\ \hline
W/o logits loss & \multicolumn{1}{l|}{99.2\%} & \multicolumn{1}{l|}{0.678} & \multicolumn{1}{l|}{0.418} & \multicolumn{1}{l|}{-0.262} & 14.9 \\ \hline
\end{tabular}
}
\end{table}

\begin{table}[t!]
\footnotesize
\centering
\caption{Quantitaive comparison with state-of-the-art}
\label{tab:scout_comparison}
\begin{tabular}{|l|l|l|l|l|}
\hline
\multicolumn{1}{|c|}{\multirow{2}{*}{VGG-16}} & \multicolumn{2}{c|}{Beginner} & \multicolumn{2}{c|}{Advanced} \\ \cline{2-5} 
\multicolumn{1}{|c|}{} & \multicolumn{1}{c|}{Recall} & \multicolumn{1}{c|}{Prec.} & Recall & Prec. \\ \hline
GradCAM \cite{selvaraju2017grad} & 0.03 & \textbf{0.80} & 0.07 & \textbf{0.41} \\ \hline
Wang \textit{et al.} \cite{wang2020scout} & 0.02 & 0.72 & 0.08 & 0.37 \\ \hline
Proposed MC CFE & \textbf{0.05} & 0.78 & \textbf{0.1} & \textbf{0.41} \\ \hline
\end{tabular}
\end{table}

\subsection{Comparison with the state-of-the-art} \label{sub-compSOTA}
\textcolor{black}{ Quantitative evaluation and comparison of state-of-the-art counterfactual explanation models is challenging as the ground truths are unavailable. Recent works have mostly provided qualitative or human-based evaluations \cite{pmlr-v97-goyal19a, akula2020cocox}. To address the lack of quantitative comparisons, Wang \textit{et al.} \cite{wang2020scout} proposed a way to synthetically generate ground truths based on part and attribute annotations present in the CUB dataset \cite{WahCUB_200_2011}. To synthesize the ground truths, the authors identified a list of parts that distinguish each class pair based on the distribution of attributes present in images. The authors used the recall and precision metrics to quantify the performance of their counterfactual explanations. For a given pair of predicted and counterfactual classes for an input image, recall is computed by finding the ratio of ground truth parts lying within the explanation region and the total number of ground truth parts \cite{wang2020scout}. Similarly, precision is computed by finding the ratio of ground truth parts and the total number of annotated parts in the explanation region. Based on these metrics, we present a comparison of the proposed MC CFE model with Wang \textit{et al.}'s \cite{wang2020scout} counterfactual explanations for images belonging to five chosen classes shown in Table \ref{tab:scout_comparison}, for the same pre-trained VGG-16 model. We followed a similar evaluation strategy as described by Wang \textit{et al.} \cite{wang2020scout} to explain model predictions with respect to the counterfactual classes chosen to simulate beginner and advanced users. For beginner users, the counterfactual class is chosen randomly, whereas, for advanced users, the counterfactual class is chosen as the top-2 predicted class. We have also included a comparison with the default GradCAM \cite{selvaraju2017grad} based visual explanation method. \textcolor{black}{Beginner user explanations are easier to generate as there is a large difference between the predicted and random counterfactual classes. Due to this, there are more distinct ground truth parts, leading to lower recall but higher precision values.} Whereas, for advanced users, the explanations are harder since the predicted and the counterfactual class pairs are closely related to each other leading to fewer distinguishable parts. This leads to higher recall and lower precision scores. From Table \ref{tab:scout_comparison}, it can be seen that the recall and precision scores of the proposed CFE method are consistently better than Wang \textit{et al.}'s \cite{wang2020scout} method. A higher recall score indicates that the proposed explanation can identify more distinct ground truth parts that separate the predicted and counterfactual classes compared to other methods. \textcolor{black}{The GradCAM visual explanation method produces higher precision scores because it generally produces a larger explanation area that encompasses more parts in the image. This leads to a higher ratio of ground truth parts in the explanation region.} One drawback of this comparison method developed by Wang \textit{et al.} \cite{wang2020scout} is that the synthesized ground truths for different class pairs are not always accurate. For example, the distinct ground truth parts for the class pair of ``Red-winged blackbird" and ``Bronzed cowbird" (shown in Fig. \ref{fig:results_1}) as generated by \cite{wang2020scout} consists of just the different colored eyes of the birds. However, these classes have additional distinguishing features of wing color and pattern, which have not been captured by \cite{wang2020scout}'s method. The same problem is true for a few other classes as well. There is a need for developing clear and robust evaluation metrics for counterfactual explanations that the research community should look into in the future.}

\subsubsection{Discussion on other related works}
\textcolor{black}{This section discusses the quality of explanations provided by the proposed CFE method and other existing works with a similar objective.} Goyal \textit{et al.} \cite{pmlr-v97-goyal19a} proposed a counterfactual explanation method for images that repetitively alters the region in the input image until the model predicts class $c'$ instead of $c$. This method can identify the important features relevant to different classes. However, it provides visual explanations only and does not explore the internal workings of the DCNN model. \textcolor{black}{Our CFE model, on the other hand, identifies the filters and corresponding high-level concepts associated with different classes to give contrastive and counterfactual explanations.} These filters are modified to alter model decisions and establish an understanding of the internal working of the model, thus improving transparency and reliability.
Dhurandhar \textit{et al.} \cite{dhurandhar2018explanations} proposed a contrastive explanation method that identifies minimally sufficient features and those features whose absence is essential to maintain the original decision. The objective of this work is similar to ours, but again it operates on pixel level and does not identify high-level semantically meaningful features, unlike our method.
Akula \textit{et al.} \cite{akula2020cocox} also proposed a counterfactual explanation methodology that identifies meaningful concepts using super-pixels that were added or removed from images to provide explanations. Although this method provides valuable explanations, they are only based on globally identified concepts instead of locally identifying features from the image that may not be actively used in the decision-making process. 
\textcolor{black}{The proposed CFE model, on the other hand, is a predictive model that locally explains each image based on counterfactual and contrastive explanations.} We identify the most important features or concepts (filters) that are actively being used in the decision-making process for an image that separates the model's decisions between classifying the image either to the inferred class or to the target class. In another recent study, Ghorbani \textit{et al.} \cite{ghorbani2020neuron} proposed a method that quantifies the contribution of each filter in a DCNN towards the model's performance. \textcolor{black}{This method identifies critical filters that considerably lower the model's performance when removed from the model.} Although this method probes the internal working of a DCNN, it does not provide contrastive or counterfactual explanations. \textcolor{black}{The proposed CFE model provides such explanations and identifies the critical filters that are important towards different classes.} We have shown that the removal of these filters considerably lowers the model's class recall without affecting the overall model performance.

\section{Conclusion}
\label{sec:five}

\textcolor{black}{This paper introduced a novel method for explaining deep CNN models based on counterfactual and contrastive explanations.} The proposed method probes the internal workings of DCNNs to predict the most important filters that affect a pre-trained model's predictions in two ways. \textcolor{black}{Firstly, we identified the minimum correct (MC) filters for a given input image.} If they were the only ones active, the model would still classify the input to the original inferred class. Secondly, we identified the minimum incorrect (MI) filters. If they were altered in a certain way, the model would have classified the image to some alter class instead of the original inferred class. We showed that these filters represent the critical features or concepts that the DCNN model learns to detect and rely on for making decisions. We discussed the effect of enabling or disabling these filters/features and also discussed misclassification detection as an application of the proposed methodology. \textcolor{black}{With these explanations, we showed reasoning behind the model decisions and improved model understanding and reliability.}

In the future, we intend to improve the evaluation metrics for better performance comparison of counterfactual explanation models. 
\textcolor{black}{Adversarial attack detection, model debugging, and machine teaching are possible applications of the proposed methodology that can be explored in the future.} 





\bibliographystyle{plainnat}
\bibliography{bibliography}






\end{document}